\documentclass{article} 
\usepackage{iclr2026_conference,times}

\usepackage{amsmath,amsfonts,bm}

\def\eqref#1{equation~\ref{#1}}

\def\1{\bm{1}}

\DeclareMathAlphabet{\mathsfit}{\encodingdefault}{\sfdefault}{m}{sl}
\SetMathAlphabet{\mathsfit}{bold}{\encodingdefault}{\sfdefault}{bx}{n}

\def\sR{{\mathbb{R}}}

\newcommand{\SO}{\mathbb{SO}\!\left(3\right)}
\newcommand{\SE}{\mathbb{SE}\!\left(3\right)}
\newcommand{\SIM}{\mathbb{S}\mathrm{im}\!\left(3\right)}

\newcommand{\Rcam}[1]{{R^{\mathrm{cam}}_{#1}}}
\newcommand{\tcam}[1]{{t^{\mathrm{cam}}_{#1}}}

\newcommand{\TT}{\mathcal{T}}
\newcommand{\OO}{\mathcal{O}}
\newcommand{\RR}{\mathcal{R}}

\newcommand{\pcan}{\boldsymbol{p}_{\mathrm{c}}}

\newcommand{\pcam}[1]{\boldsymbol{p}^{\mathrm{cam}}_{#1}}

\newcommand{\pp}{\boldsymbol{p}}

\usepackage{multirow}
\usepackage{hyperref}
\usepackage{xcolor}
\usepackage{titlesec}

\titlespacing*{\paragraph}
  {0pt}      
  {0.0em}    
  {0.5em}    

\hypersetup{
    colorlinks=false,
    citebordercolor={0 1 0},
    urlbordercolor={1 1 1}, 
}

\newcommand{\magentaurl}[1]{%
    \begingroup
    \hypersetup{pdfborder={0 0 0}}%
    \href{#1}{\textcolor{magenta}{\nolinkurl{#1}}}%
    \endgroup
}

\usepackage{url}
\usepackage{cleveref}
\usepackage{booktabs}
\usepackage{graphicx}
\usepackage{wrapfig}
\usepackage{caption}
\usepackage{tabularx}
\usepackage{adjustbox}

\title{AgentSTAR: Agentic Shape Tracking and Reconstruction from Monocular Videos}

\author{
\begin{tabular}{@{}l@{}}
{\bfseries Kirill Mazur, Nikita Karaev, Matthew Chang,}\\
{\bfseries Jitendra Malik, Nur Muhammad ``Mahi'' Shafiullah}\\[0.3em]
{\normalfont Amazon FAR (Frontier AI and Robotics)}\\
{\normalfont\ttfamily \{makezur,nikaraev\}@amazon.co.uk}\\
{\normalfont\ttfamily \{drmchang,jtnmalik,notmahi\}@amazon.com}
\end{tabular}
\\[0.5em]
\\
{\normalfont\makebox[\textwidth][c]{\magentaurl{https://agenticstar.github.io}}}
}

\iclrfinalcopy 
\begin{document}
\maketitle
\fancyhead[L]{}
\vspace{-2.5em}
\begin{figure}[h!]
    \centering
    \includegraphics[width=\linewidth]{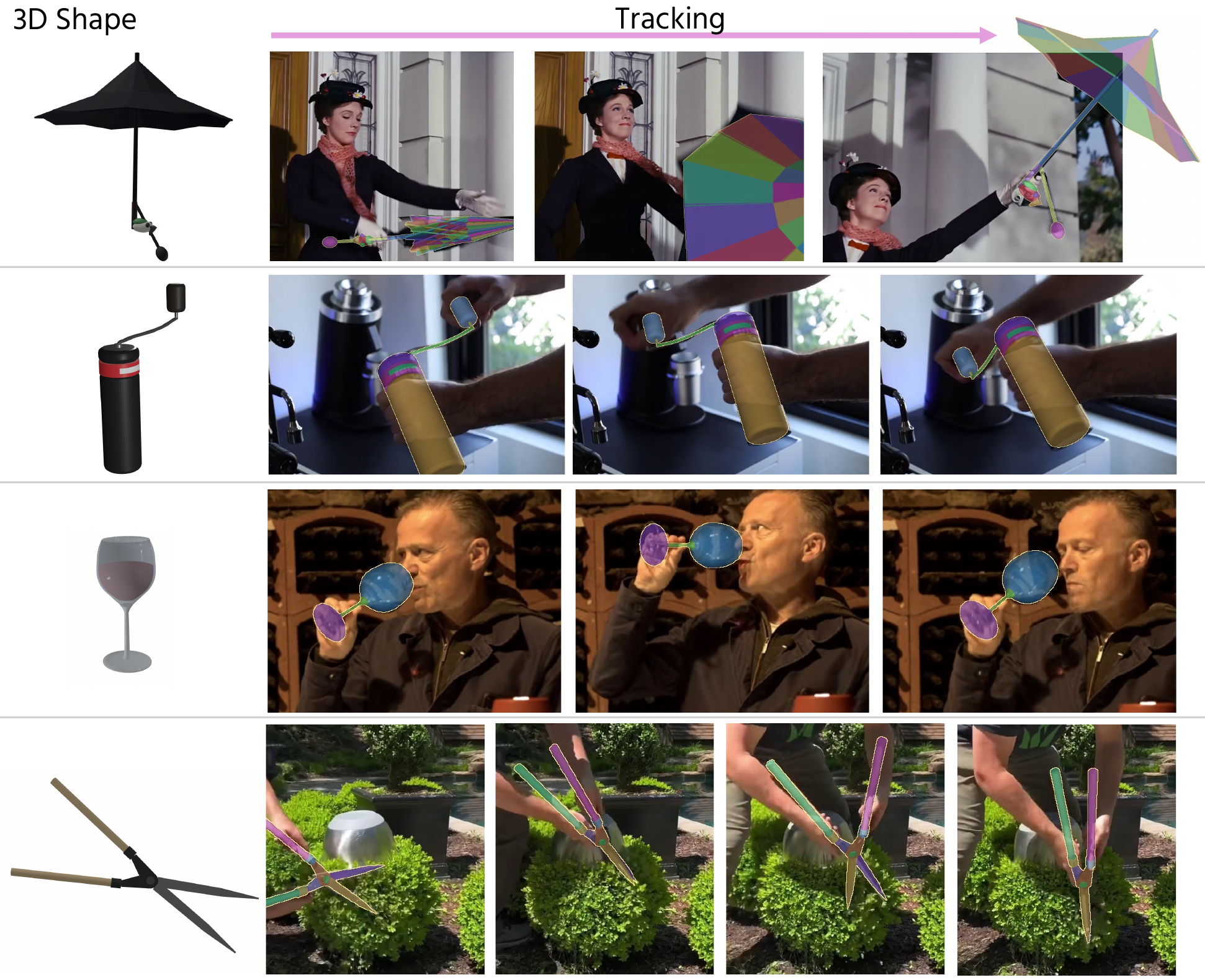}
    \vspace{-1.5em}
    \caption{
    \textbf{Agentic shape tracking and reconstruction.}
Our method jointly reconstructs object geometry and kinematics (\textbf{left}) from monocular videos and tracks the resulting 3D model through time (\textbf{right}).
It recovers complex articulation (first and fourth rows), preserves fine geometric details (second row), and tracks transparent objects (third row). Consistent part colours across frames visualise temporal correspondences and tracking quality.
    }
    \label{fig:placeholder}
\end{figure}

\begin{abstract}
 
In this work, we present a method for shape reconstruction and tracking from video via agentic analysis-by-synthesis. Unlike prior methods which first estimate dense pixel correspondences and then recover object motion from them, our method infers a structured 3D object model, including its geometry and kinematic structure, and uses this model to optimise object track estimates over time.
In our optimisation loop, a Vision-Language Model (VLM) agent iteratively refines shape or generalised pose through a render-and-compare loop, combining coarse visual reasoning with numerical pose optimisation for precise state estimation. This structured formulation enables our method to track through large motion, articulation, and severe occlusion without relying on pixel-matching objectives. Quantitatively, on ARCTIC, our method substantially outperforms state-of-the-art 3D point-tracking baselines for articulated objects, and on HOT3D it outperforms all evaluated rigid-object tracking baselines.
 
\end{abstract}

\section{Introduction}

Should we first recover low-level visual evidence, such as scene flow or point trajectories, and then infer the object that generated it? Or should we first recognise the object and its structure, and use this structured representation to recover its state over time? Most previous work on articulated object reconstruction~\citep{liu2023reart, jiayi2023paris, zhao2025real2code, peng2025itaco, delitzas2026funrec} approaches the problem in a bottom up way, building on general scene-flow and correspondence methods. These methods reconstruct the 3D scene and estimate either 2D correspondence~\citep{teed2020raft, karaev23cotracker, harley2025alltracker} or 3D point trajectories~\citep{d4rt, vdpm, st4rtrack2025, xiao2025spatialtracker} over time. 
While this provides a powerful general-purpose representation of observed motion, it comes with two key limitations. First, dense correspondence estimation and tracking are themselves challenging under the occlusions and limited visual overlap typical to dynamic object interactions. 
Second, the resulting representation remains largely unstructured: point clouds and trajectories describe where visual evidence moves, but not the underlying objects, their parts, joints, or state variables. 
Such structure must therefore be inferred only after reconstruction, which becomes brittle when the recovered geometry and trajectories are imperfect. This is particularly limiting for downstream applications such as robotics, which require explicit object models that can be instantiated and manipulated in simulation.

Rather than first asking where every observed point moved, we directly ask which \emph{structured 3D object and state sequence could have generated the video}. 
We therefore adopt an analysis-by-synthesis approach for dynamic object reconstruction and tracking. Our representation models the object's geometry, kinematic structure, and generalised pose over time, comprising its 6-DOF base pose and kinematic states. Once the object's structure is known, its permissible configurations are strongly constrained, transforming dense point-wise motion estimation into a compact pose-estimation problem over base pose and articulation. The resulting states are also interpretable and semantically meaningful.

Realising this top-down approach requires a model that can infer object structure and then use that structure to inform pose estimation. Much of an object’s motion is determined not by visual evidence alone, but by its internal mechanism. For example, when a person closes a book, its pages become occluded and can no longer be visually tracked, yet their possible motion remains strongly constrained by the structure of the book. To exploit knowledge of the object’s mechanism, tracking should be done by the same entity that inferred the shape.

Large vision-language models (VLMs), which serve as the backbone of modern coding
agents, now exhibit a high level of generalisation and are extremely good at
providing coarse 3D shape estimates~\citep{yin2026viga, zhou2026articraft} or detecting gross spatial inconsistencies, making them natural candidates for this task. On their own, however, they are poor state
estimators: a VLM can reliably judge whether an object is \textit{roughly} posed
correctly (e.g., that it is upside down), but cannot recover its pose to within a few degrees. Traditional test-time optimisation methods such as Structure-From-Motion (SfM), on
the other hand, are often precise but brittle.

We show how to pair VLM agents with numerical optimisation tools to form an \textit{agentic optimisation loop} for joint shape and state
estimation, iteratively recovering the shape, joints, and pose of dynamic objects. The key difference from existing work is that the VLM makes gradient-free optimisation tractable, by narrowing the search to a promising optimisation basin and rejecting spurious local minima, which is often the hardest
part of the problem. Once the correct basin is identified, the compact parameter space of shape parameters, joint axes and their configurations makes
refinement tractable.

To the best of our knowledge, ours is the first work to apply VLM agents to joint shape and motion reconstruction of both
rigid and articulated objects from casually captured monocular videos. We propose a novel agentic render-and-compare loop, and demonstrate that our method outperforms 3D point tracking baselines for articulated object reconstruction on ARCTIC~\citep{fan2023arctic} and existing articulated object reconstruction methods on   iTACO~\citep{peng2025itaco}, as well as rigid object tracking systems on the challenging HOT3D~\citep{banerjee2025hot3d} dataset. Crucially, because we model the cause of object motion rather than its visual evidence, the proposed method can reconstruct and recover motion in cases that are beyond the reach of prior methods --- for example, tracking transparent objects such as glass, as well as severely occluded or only partially visible objects.

\section{Related Work}

\paragraph{3D as Code.}
Structured code representations have been explored for vector graphics, CAD, and meshes~\citep{carlier2020deepsvg,Wu2021deepcad,nash2020polygen,siddiqui2023meshgpt}. SceneScript~\citep{avetisyan2024scenescript} brought this paradigm to perception, predicting structured 3D scene primitives from visual observations and SfM information. More recently, LLMs, VLMs, and agentic systems have been used for code-based 3D generation and inverse graphics~\citep{kulits2024re,gu2025blendergym,sun20253d,yin2026viga}. Code-based representations are particularly natural for articulated objects, whose geometry and kinematics are hierarchical. Starting with Real2Code~\citep{zhao2025real2code}, subsequent systems~\citep{le2024articulate,zhou2026articraft} reconstruct or generate articulated models in code. These approaches target model acquisition or asset generation rather than reconstruction and tracking through video.

\paragraph{Articulated Object Reconstruction.} 
Early work recovered articulated structure by factorising tracked trajectories into rigidly moving parts and their kinematic relations~\citep{kanadefactorise98,yan2008factorization,tresadern2005articulated}. Modern bottom-up methods recover part geometry and motion from multi-view consistency, neural fields, or dense correspondences~\citep{jiayi2023paris,liu2023reart,deng2024articulate,kerr2024rsrd,Mazur:etal:CVPR2026, delitzas2026funrec}. Learning-based methods directly predict articulated geometry or kinematics from static or sparse observations~\citep{jiang2022ditto,heppert2023carto,li2025art,li2026particulate}. More recent approaches strengthen these priors using retrieval, generative models, and VLMs~\citep{urdformer_2024,le2024articulate,jiayi2024singapo,zhao2025real2code,li2025urdfanything}. While these methods can infer plausible kinematic structure from limited observations, they generally reconstruct an articulated model rather than track its articulation through a video.
\paragraph{Object and Point Tracking.} 
Tracking methods estimate motion either as point trajectories or as coherent object motion. Point trackers estimate 2D~\citep{doersch2022tap,doersch2023tapir,karaev23cotracker,harley2025alltracker} or 3D~\citep{koppula2024tapvid,SpatialTracker,xiao2025spatialtracker,vdpm,d4rt} trajectories, but provide no explicit object structure. Rigid object trackers instead impose a shared 6-DoF motion and are either model-based or model-free. Model-based methods such as FoundationPose~\citep{foundationposewen2024} require an object model; recent systems obtain one through 3D generation~\citep{nguyen2024gigaPose,lee2025any6d}, including SAM3D~\citep{sam3dteam2025sam3d3dfyimages} adapted for monocular tracking~\citep{paliwal2026doasido}. Model-free methods~\citep{sun2022onepose,wen2023bundlesdf,taher2026kv} jointly recover structure and pose from observations. Overall, point trackers provide flexible but unstructured trajectories, whereas rigid object trackers provide structure but cannot represent articulation.

\section{Method}

\begin{figure}[h]
    \centering
    \includegraphics[width=\linewidth]{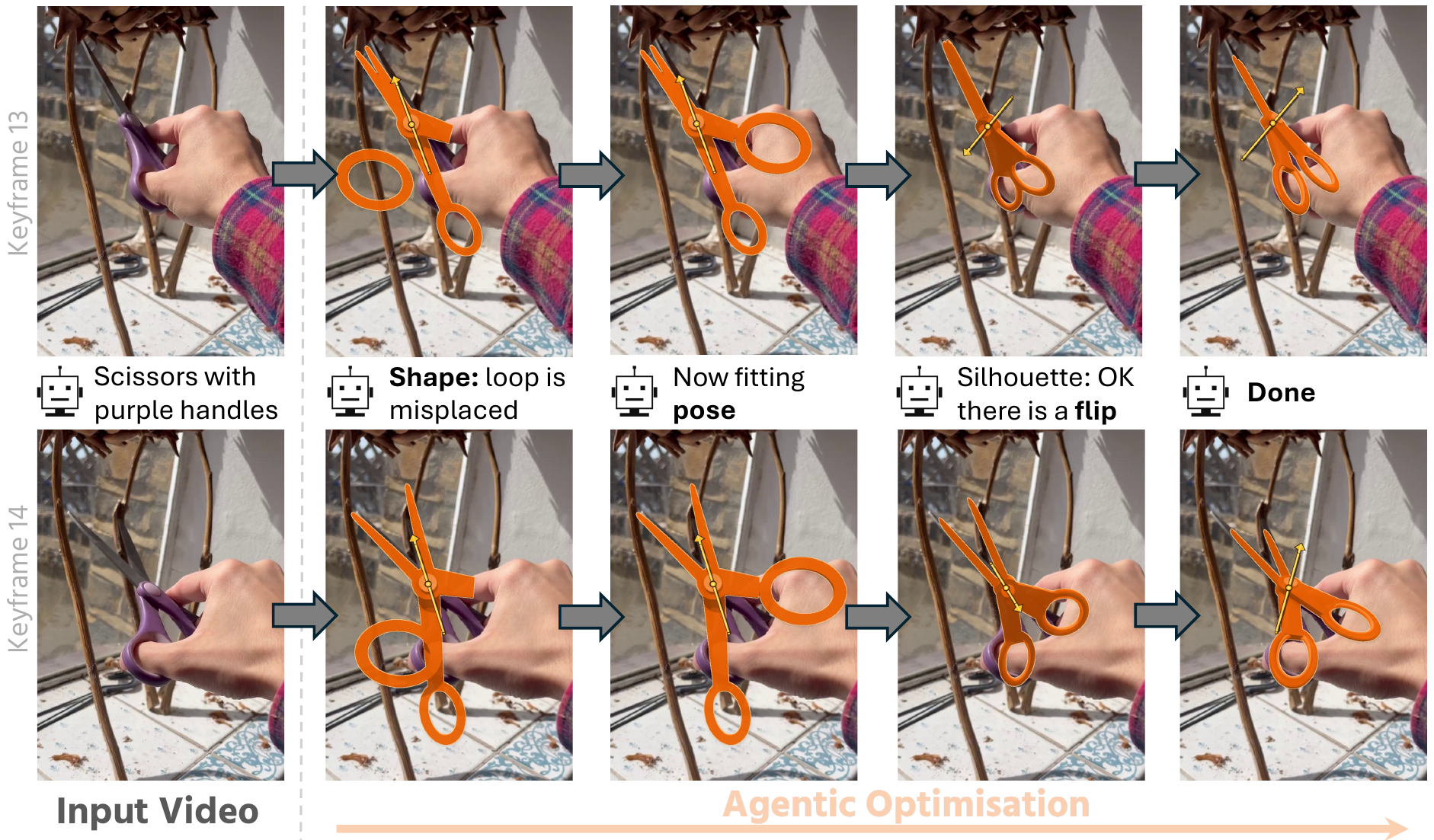}
    \caption{\textbf{Method}. Given an input video \textbf{(left)}, our agentic optimisation loop iteratively refines the shape and pose of the observed object \textbf{(right)}. The middle panels summarise the agent's reasoning throughout the optimisation process. Our harness provides the VLM with numerical optimisation tools for precise pose fitting, together with temporal diagnostics for assessing consistency across keyframes. The final output is a reconstructed object model together with its estimated generalised pose at every observed keyframe.}
    \label{fig:method}
\end{figure}

\paragraph{Task formulation.} Our method takes as input a set of observed video frames $\{I_i \in \sR^{H \times W \times 3}\}$, \textit{optionally} coupled with depth maps $\{D_i \in \sR^{H \times W} \}$, and binary masks $\{ M_i \in \{0, 1\}^{H \times W } \}$ for the target object. Our method can also take in hand segmentation masks $\{ H_i \in \{0, 1\}^{H \times W } \}$, as they often occlude the object of interest. The masks are typically extracted by a video-segmentation model, such as SAM3~\citep{carion2025sam3segmentconcepts}, or provided directly.

\paragraph{Goal.} Our goal is to build a canonical object model $\mathcal{O}$, represented as code and shared across all observed frames, together with its poses and kinematic parameters $\TT_i$ for every observed video frame $I_i$. We use an object-to-camera convention for poses and forward kinematics, so a generalised pose should map the object to its coordinates in the target camera.

\paragraph{Conventions.}  We assume a pinhole camera model with known calibration $K_i$ and known camera pose extrinsics $(\Rcam{i},\tcam{i})$ estimated by an external SLAM system or a feed-forward reconstruction system~\citep{wang2025vggt, wang2025pi3}. Given object model $\mathcal{O}$ and its pose $\TT_i$ we denote its render into the camera $i$ as $\RR(\OO, \TT_i)$. Its silhouette render is denoted as $\RR_s(\OO, \TT_i)$

\subsection{Overview}
Recent large models have been trained on a vast amount of data and exhibit a high level of semantic knowledge and recognition capability, including the ability to recognise shape, kinematic structure, and approximate object orientation~\citep{zhou2026articraft}. However, these models struggle with precise quantity estimation. Estimating continuous quantities, such as object pose, has always been the strength of iterative, optimisation-based methods, such as bundle adjustment~\citep{Triggs:etal:VISALG1999}. Guided by these observations, we design an \emph{agentic render-and-compare} optimisation loop that allows VLMs to iteratively refine both the shape and pose of the object. Akin to regular optimisation methods, our core loop is iterative. Each iteration is allowed to either modify the shape or pose only. The exact scheduling decision is left to an agent. 

\paragraph{Shape step.} Shape and kinematics editing is done by a coding agent, conditioned on the past observation history, such as observations made during pose steps. All shape coding is confined to a single scene.py script, where geometry is freeform and can represent arbitrary topologies, whereas pose definition should follow the representation conventions: all pose-related fields are stored under predefined variable names in the script, so they can be automatically decoupled from the shape definition and updated independently. Diagnostic rendering tools allow the agent to inspect the resulting object from both observed and novel viewpoints.

\paragraph{Pose step.} 
For a pose iteration, the object model $\mathcal{O}$ is frozen and rendered under its
current pose estimates $\{ \TT_i \}$ at the target keyframes. Rather than directly
predicting precise continuous pose updates, the agent specifies interpretable
directions or search regions of the pose space to explore. Our pose optimisation
tool~(\ref{subsec:pose}) searches these regions numerically, renders and scores candidate poses. The top scoring candidates are then returned back to the agent for visual inspection. 
Pose refinement proceeds iteratively: the agent invokes the pose optimisation tool, visually inspects the returned candidates, and uses the results to specify new search directions for subsequent iterations. Since the shape is shared and fixed during these iterations, pose refinement can be parallelised across the sequence.

For video, the agent has to run a mandatory temporal diagnostic tool~(\ref{sec:temporal}) that identifies implausible discontinuities in the estimated trajectory. This allows temporal smoothing to be applied selectively rather than imposed as a fixed prior.

\subsection{Scoring Objective}
\label{sec:scoring}
Inspired by the shape-from-silhouette~\citep{szeliski1993rapid} line of work, we set Intersection-over-Union (IoU) of our renders' silhouettes $\RR_s(\OO, \TT)$ with the target object mask $M_i$ as the main numerical score $\mathcal{S}(\OO, \TT)$ for the agent. Although IoU is an inherently local signal and is prone to many local minima, coupling it with a VLM's ability to reason ``approximately'' helps the agent escape these minima. We focus on dynamic objects and assume humans are manipulating them. We extract hand segmentation masks $H_i$ and exclude them from IoU, as hands often severely occlude the object of interest. Formally, 
\begin{equation}
    \label{eq:score}
    \mathcal{S}(\OO, \TT)  \; \colon= \operatorname{IoU} (\RR_s(\OO, \TT) \setminus H_i ; M_i \setminus H_i )
\end{equation}

Incorporating other signals, such as pixel matching~\citep{teed2020raft, mast3r_eccv24, edstedt2024roma}, might also be viable, although the agentic optimisation loop should then reason about the failure cases of the possibly noisy signal. We therefore do not use such signals. We also provide a variant of our system where additional supervision (score term with a blending coefficient $\alpha$) comes from depth signal. Unless stated otherwise, our method does \emph{not} employ depth supervision.

\subsection{Object and State Representation}
\label{sec:conventions}
\paragraph{Shape and Geometry.} Our shape and its kinematic structure $\mathcal{O}$ are represented in the form of Python code. Geometry is expressed via Blender shape primitives. The joint kinematic structure, its limits, and the joint positions are also expressed this way, which allows modelling arbitrary kinematic structures and permits natural updates to the shape or state space of an object of interest in the form of code diffs.

\paragraph{Pose Representation.} 
For an articulated object, we define the \textit{generalised pose}
$\mathcal{T}_i$ as the 6-DoF pose of its base together with the states of
all articulation joints:
\begin{equation}
    \mathcal{T}_i =
    \left(T_i, j_{i,1}, \ldots, j_{i,n}\right),
    \qquad
    T_i=(R_i,t_i)\in \SE,
    \qquad
    j_{i,k}\in \mathbb{R}^1
\end{equation}
Each joint state $j_{i,k}\in\mathbb{R}^1$ is constrained by its corresponding
motion limits. Throughout the paper, we use \emph{pose} to refer to this
generalised pose unless stated otherwise.

The object has a single scale $s$ shared across all frames, and we estimate
object-to-camera poses. We store rotations as unit quaternions and
translations as vectors. Given a canonical-frame point $\pcan$ after
forward kinematics, its position in camera $i$ is $\pp_i = s R_i \pcan + t_i$.
\paragraph{Pose Updates.}
To make rotational corrections interpretable to the VLM, we distinguish between camera- and object-centric updates. Because rotations do not commute, these correspond to left- and right-multiplication of the current rotation, respectively, and therefore to rotations expressed in different coordinate frames. Object-centric updates are particularly useful for large transformations such as symmetry flips. Assuming for clarity that the object centre after forward kinematics is at the canonical origin, the two update conventions are:
\begin{equation}
\begin{aligned}
\pcam{i} &= s(\Delta R)R_i \pcan + t_i + \Delta t
&&\text{(camera-centric)}\\
\pcam{i} &= sR_i(\Delta R) \pcan + t_i
&&\text{(object-centric)}
\end{aligned}
\label{eq:pose_updates}
\end{equation}

\subsection{VLM-Guided Pose Optimisation}
\begin{figure}[t]
    \centering
    \includegraphics[width=\linewidth]{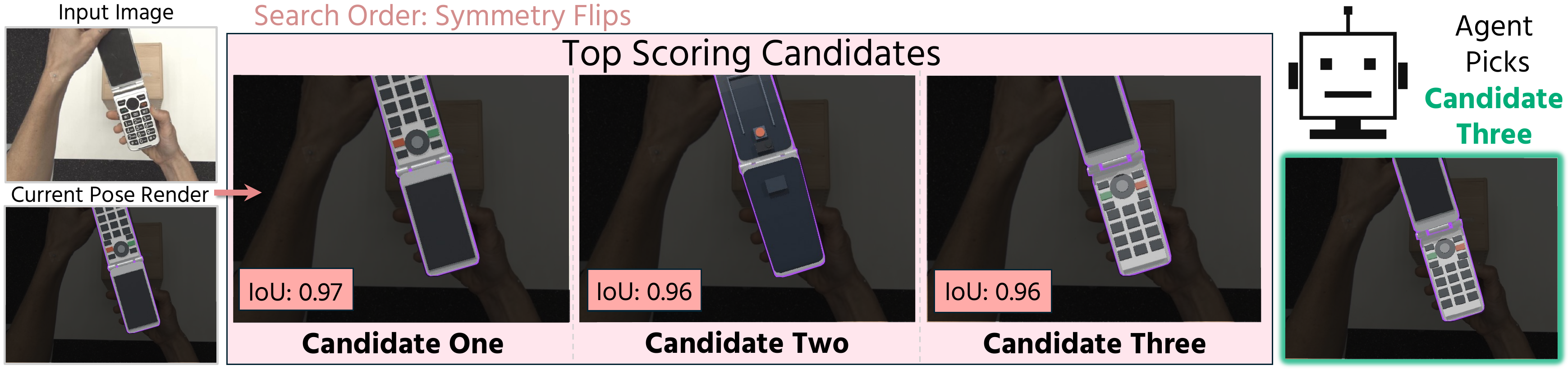}
    \caption{\textbf{VLM-guided pose optimisation.}
Given the current pose estimate, a VLM agent defines a bounded search subspace, such as symmetry flips around a chosen axis.
A numerical optimiser searches this subspace and returns the highest-scoring pose candidates.
Because the silhouette-based objective is prone to spurious local optima, the VLM visually inspects the top candidates and selects the best pose, which becomes the starting point for the next optimisation iteration.}
    \label{fig:VLM_guide}
\end{figure}

\label{subsec:pose}

Classical pose optimisation methods can yield precise estimates, but our scoring objective
$\mathcal{S}(\OO,\TT)$ provides an inherently local signal and is therefore sensitive
to initialisation and prone to spurious local optima. In contrast, modern VLM agents
are effective at coarse visual reasoning, but less suited to precise continuous
estimation. We combine these complementary strengths in a \emph{VLM-guided pose
optimisation loop}, where the VLM identifies a promising search region and a numerical
optimiser performs precise refinement within it.

Given the current pose $\TT$, the agent specifies a bounded search region, e.g. ``vary yaw from $-15^{\circ}$ to $15^{\circ}$ and the hinge joint from $24^{\circ}$ to $64^{\circ}$.'' A gradient-free optimiser~\citep{Storn1997} is then executed within this region, producing the $K$ highest-scoring pose candidates
$\{\TT_{\mathrm{order}}^i\}_{i=1}^{K}$.
Their corresponding renders and numerical scores $\left\{
    \RR(\OO,\TT_{\mathrm{order}}^i),
    \mathcal{S}(\OO,\TT_{\mathrm{order}}^i)
    \right\}_{i=1}^{K}$ are returned to the agent for visual inspection. The final selection is made by the VLM, which chooses the \emph{visually} best candidate among these high-scoring solutions. Consequently, the selected pose need not be the numerical optimum under $\mathcal{S}$. The selected candidate becomes the starting pose for the next optimisation iteration.

\paragraph{Pose Search Space.}
The agent specifies bounded search intervals over interpretable pose variables,
including yaw, pitch, roll, image-plane translation, depth, and articulation states.
We describe the pose-update parameterisation and transformation conventions in~\cref{sec:conventions}.

\subsection{Sequence-level Pose Estimation}
When fitting an object model to a video, estimating poses independently across frames can produce temporally inconsistent trajectories. For example, for symmetric objects, adjacent frames may converge to different symmetry-equivalent poses, resulting in discontinuous 3D point trajectories. Classical state-estimation methods~\citep{Dellaert:Kaess:Foundations2017} often address this with temporal smoothness factors between consecutive estimates. However, a fixed smoothness prior can suppress genuine rapid motion, which is common in real-world videos. We therefore use \emph{agentic smoothing}: temporal inconsistencies are reported to the agent as diagnostics, while the agent can retain rapid motion when it is supported by the visual observations.

\paragraph{Temporal Residuals.}
\label{sec:temporal}

Given a sequence of generalised poses $\{\TT_0,\TT_1,\ldots,\TT_n\}$, we compute temporal residuals for the object's 6-DoF base pose and its joints. Specifically, we measure velocities and accelerations component-wise rather than defining a single metric over the full configuration space. For one-dimensional joints, these quantities are given by first- and second-order differences.

For the monocular setting with known camera poses, the dynamic object retains an independent scale gauge. We therefore reparameterise its translation as $t_i = s\tau_i$:
\begin{equation}
    \label{eq:x_world}
    \pp^{world}_i
    = s(\Rcam{i} R_i)\pp
    + \left(s\,\Rcam{i}\tau_i + \tcam{i}\right).
\end{equation}
For temporal residuals, we compare the camera-motion-compensated rotation
$\Rcam{i}R_i$ and the scale-normalised translation $\Rcam{i}\tau_i$,
expressed in object units. We omit $\tcam{i}$ because the camera trajectory's
translation gauge need not be compatible with the scale of the dynamic object reconstruction.

\section{Experiments}
\subsection{Experimental Details}
Unless stated otherwise, we use the Codex harness with our tool suite and GPT-5.6 Sol at medium reasoning effort. We employed the same set of tooling and prompts for all experiments, unless stated otherwise. We budget 10 hours for each optimisation loop, however some agents can report termination earlier. 

\subsection{Articulated Object Reconstruction: ARCTIC}

\begin{table}[h]
    \centering
    \caption{\textbf{Performance on ARCTIC.}
    \textbf{(Left)} 3D tracking quality compared with state-of-the-art 3D point
    trackers: V-DPM~\citep{vdpm}, OpenD4RT~\citep{d4rt}, and
    SpatialTrackerV2~\citep{xiao2025spatialtracker}.
    We evaluate points queried in the first keyframe and track their 3D trajectories
    across all subsequent keyframes.
    \textbf{(Right)} Geometry reconstruction quality compared with V-DPM,
    the best-performing tracking baseline in the left panel.
    Errors are reported in cm.}
    \label{tab:arctic}
    \begin{minipage}[t]{0.66\linewidth}
        \centering
         \fontsize{8pt}{9.5pt}\selectfont
        \begin{tabular*}{\linewidth}{
            @{\extracolsep{\fill}} lcccc @{}
        }
            \toprule
            \textbf{Metric}
            & \textbf{SpatialTrackerV2}
            & \textbf{OpenD4RT}
            & \textbf{V-DPM}
            & \textbf{Ours} \\
            \midrule
            3D EPE (cm) $\downarrow$
            & 10.79
            & 8.77
            & 7.65
            & \textbf{5.59} \\
            \bottomrule
        \end{tabular*}

        \vspace{0.3em}
        \textbf{(a) 3D Tracking}
    \end{minipage}%
    \hfill%
    \begin{minipage}[t]{0.32\linewidth}
        \centering
         \fontsize{8pt}{9.5pt}\selectfont
        \begin{tabular*}{\linewidth}{
            @{\extracolsep{\fill}} lcc @{}
        }
            \toprule
            \textbf{Metric}
            & \textbf{V-DPM}
            & \textbf{Ours} \\
            \midrule
            Chamfer (cm) $\downarrow$
            & 4.72
            & \textbf{3.36} \\
            \bottomrule
        \end{tabular*}

        \vspace{0.3em}
        \textbf{(b) Geometry}
    \end{minipage}
\end{table}

ARCTIC~\citep{fan2023arctic} contains humans interacting with articulated
objects undergoing substantial articulation and rapid motion. It provides
high-quality ground-truth geometry and motion captured using a motion-capture
rig.

For evaluation, we use the egocentric-camera sequences of subject $s1$,
yielding $24$ sequences in total. We discard the first $100$ frames of each
sequence, which typically correspond to capture setup, contain little or no
object motion, and often include severely underexposed frames. We divide the
remaining frames into chunks of $300$ frames and select every $10$th frame as
a keyframe. Reconstruction and tracking are performed on these keyframes, and
all methods are evaluated on the same set of keyframes. The predicted geometry and trajectories of all methods are recovered only up
to an unknown global scale. We resolve this scale through depth alignment;
see~\cref{ap:scale} for details.

\begin{wraptable}[11]{r}{0.40\textwidth}    \vspace{-1.5em}
    \centering
    \caption{\textbf{Harness ablation.}
    3D EPE for point tracking.}
    \label{tab:ablation}

    \vspace{-0.4em}
    \small
    \setlength{\tabcolsep}{3pt}
    \renewcommand{\arraystretch}{1.0}

    \begin{tabular}{@{}lr@{}}
        \toprule
        \textbf{Variant} & \textbf{EPE (cm)} $\downarrow$ \\
        \midrule
        No Harness                 & 11.26  \\
        IoU Objective              & 151.46 \\
        GT mesh + VLM-free opt.    & 14.60  \\
        No temporal                & 6.15   \\
        \midrule
        Ours (GPT 5.6 Sol)         & 5.59   \\
        Ours (Fable 5)             & \textbf{4.95} \\
        \bottomrule
    \end{tabular}

    \vspace{-0.8em}
\end{wraptable}

\paragraph{3D Point Tracking.}
Because these sequences are too challenging for existing articulated-object
reconstruction methods, we compare against state-of-the-art 3D point-tracking
methods. Most baselines are pixel-anchored: given a query pixel $q$ in the
first keyframe $I_0$, they estimate the 3D trajectory of the corresponding
scene point over time. Ground-truth trajectories are obtained from the
ground-truth articulated mesh poses. We evaluate 3D End-Point Error
(EPE)~\citep{liu:2019:flownet3d} for query pixels lying inside the
ground-truth object mask in the first keyframe.

Our representation, in contrast, is a structured 3D mesh and does not
necessarily contain a predicted point corresponding to every query pixel in
the ground-truth mask. We therefore establish correspondences between query
pixels and points on the predicted mesh. In the first keyframe, each query
pixel is matched to the nearest projected mesh point. The matched point is
then tracked through the sequence and used to compute 3D EPE.

\paragraph{Geometry Quality.}
We report Chamfer distance between the predicted and ground-truth geometry,
averaged across all keyframes. For our method, this is computed between the
predicted and ground-truth meshes. For V-DPM, we instead compute Chamfer
distance between its predicted 3D point cloud at each timestamp and the
ground-truth mesh.

\subsection{Rigid object tracking: HOT3D} 
\begin{figure}[h]
    \centering
    \vspace{-1em}
    \includegraphics[width=\linewidth]{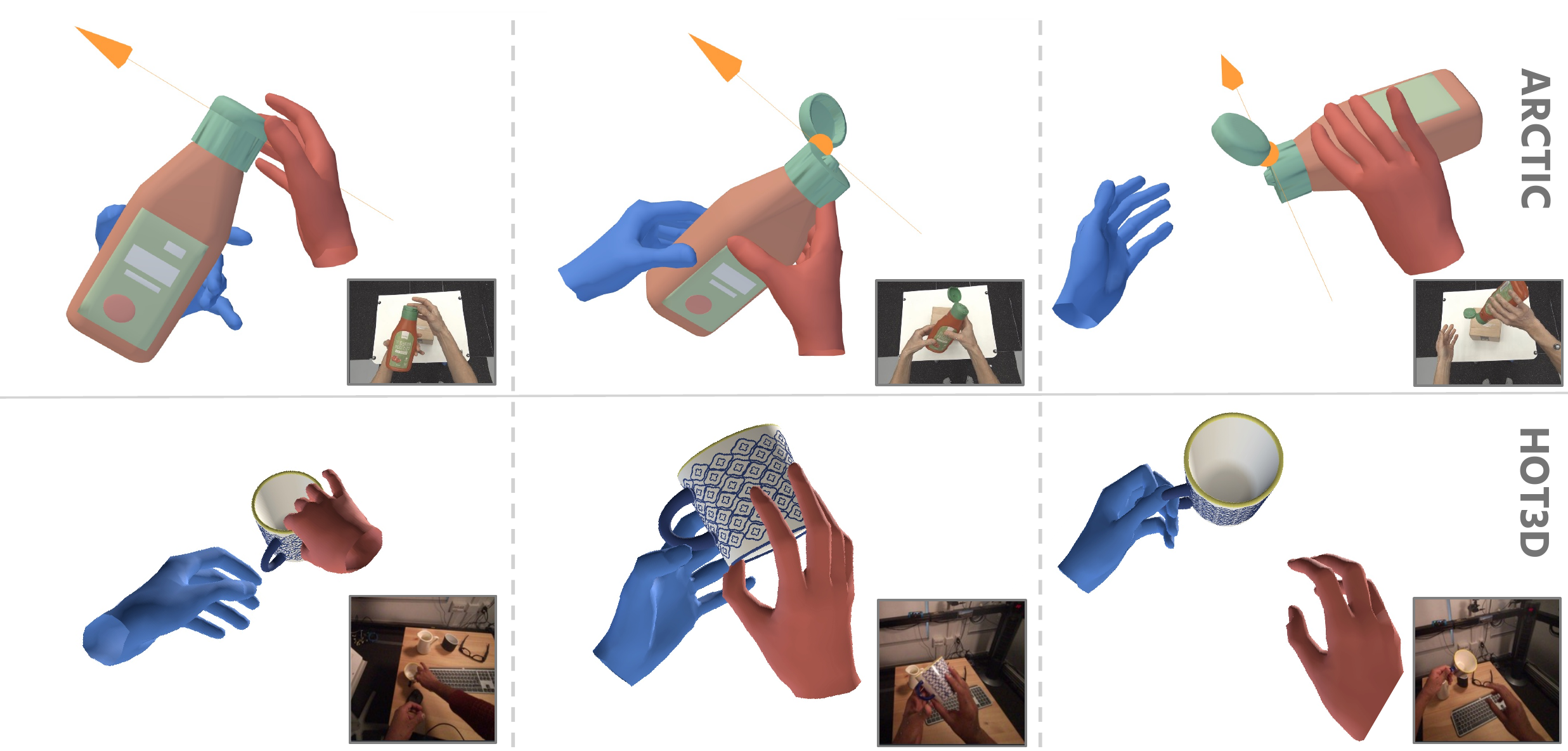}
    \caption{\textbf{Reconstructions on ARCTIC and HOT3D.} We show our reconstructions at three different timestamps alongside ground-truth hands, demonstrating the quality of our 3D pose alignment in the world frame.
\textbf{(Top)} Our method recovers fine articulations, such as the ketchup-bottle cap, while tracking large and rapid pose changes.
\textbf{(Bottom)} Our method accurately tracks objects with fine geometric details and repetitive textures.}
    \label{fig:recon_acrtic_hot3d}
    \vspace{-1em}
\end{figure}

\begin{table}[t]
\centering
\caption{\textbf{Pose Tracking on HOT3D.}
We report per-trajectory mean and median translation and rotation errors,
averaged across trajectories.
Methods that use the ground-truth CAD model are marked with *.
Our method outperforms all evaluated baselines, including those with access
to the ground-truth CAD model.
Translation errors for ProxyPose are omitted because degenerate trajectories
make the required alignment unstable.}
\label{tab:hot3d}
\begin{tabular*}{\linewidth}{@{\extracolsep{\fill}}lcccc@{}}
\toprule
\textbf{Method}
    & \multicolumn{2}{c}{\textbf{Trans. Err. (cm)} $\downarrow$}
    & \multicolumn{2}{c}{\textbf{Rot. Err. ($^\circ$)} $\downarrow$} \\
\cmidrule(lr){2-3}\cmidrule(lr){4-5}
    & \textbf{Mean} & \textbf{Median}
    & \textbf{Mean} & \textbf{Median} \\
\midrule
ProxyPose~\citep{zhang2026proxypose}
    & -- & -- & 52.7 & 43.4 \\

GigaPose~\citep{nguyen2024gigaPose}*
    & 11.25 & 8.16 & 62.8 & 50.9 \\

GigaPose* + GoTrack~\citep{nguyen2025gotrack}
    & 16.46 & 9.92 & 55.4 & 37.9 \\

FoundationPose~\citep{foundationposewen2024}* + VGGT-$\Omega$
    & 7.56 & 5.57 & 54.6 & 49.1 \\
    
SAM3D-Tracker~\citep{paliwal2026doasido}
    & 6.41 & 5.30 & 51.6 & 43.0 \\

Ours
    & \textbf{3.04} & \textbf{2.42}
    & \textbf{37.6} & \textbf{26.3} \\
\bottomrule
\end{tabular*}
\vspace{-1em}
\end{table}
\paragraph{Protocol.}
Our method also applies to model-free rigid-object 6-DoF pose estimation from monocular video. We evaluate on the challenging HOT3D~\citep{banerjee2025hot3d} dataset, which provides high-quality motion-capture ground-truth poses and contains rapid object motion. We use the validation split and retain sequences containing a single dynamic target object, resulting in $93$ sequences total. Each sequence contains $150$ frames. Our method operates on every 10th frame, yielding 15 keyframes per sequence. All methods are evaluated on exactly these keyframes. Video-tracking baselines are additionally allowed to process all $150$~frames, including the intermediate frames, but their metrics are computed only at the selected keyframes.

\paragraph{Metrics.}
Our goal is to evaluate temporally consistent object tracking rather than
frame-wise pose estimation. We therefore do not independently quotient out
object symmetries at each frame. In particular, a prediction that switches
between symmetry-equivalent poses over time is considered incorrect, since
such a switch can induce a different 3D point trajectory.

For model-free methods, the predicted canonical object frame is arbitrary,
giving rise to a global $\operatorname{Sim}(3)$ gauge ambiguity. We resolve this ambiguity by estimating a single global rotation and translation between the predicted and ground-truth trajectories, which are then fixed for
the entire sequence. For methods that also reconstruct an object (ours and SAM3D-Tracker), we estimate a single global scale factor separately by aligning rendered object depth with ground-truth depth. The same scale-alignment procedure is applied to all methods without metric-depth input. Further details
are provided in the appendix.

After alignment, we compute per-frame translation and rotation errors.
We compute the mean and median error within each trajectory, and report
their averages across the evaluation set.

\paragraph{Results.} 

As shown in~\cref{tab:hot3d}, our method outperforms all evaluated baselines
in both translation and rotation error. Rotation errors remain relatively
high for all methods, reflecting the particularly rapid motion in HOT3D:
object orientation can change by as much as $180^\circ$ between consecutive
evaluation keyframes. This regime is substantially more challenging than
rigid-object tracking settings dominated by smooth inter-frame motion.

\begin{table*}[h]
    \centering
    \caption{\textbf{Results on the iTACO benchmark.}
    We evaluate our method on the simulated RGB-D iTACO benchmark against Articulate-Anything~\citep{le2024articulate}, Robot See Robot Do~\citep{kerr2024rsrd}, and iTACO~\citep{peng2025itaco}.
Following iTACO, we report mean $\pm$ standard deviation for each metric. Our method performs competitively on geometry while substantially
outperforming the baselines on kinematic estimation.
    }
    \small
    \setlength{\tabcolsep}{5.5pt}
    \renewcommand{\arraystretch}{1.08}
    \begin{tabular}{llcccc}
        \toprule
        & Metric
        & Articulate-Anything
        & Robot See Robot Do
        & iTACO
        & \textbf{Ours} \\
        \midrule

        \multirow{3}{*}{Geometry}
        & CD-w ($m^2$) $\downarrow$
        & $0.11 \pm 0.22$
        & $3.39 \pm 21.50$
        & $\mathbf{0.01 \pm 0.01}$
        & $0.02 \pm 0.03$ \\

        & CD-m ($m^2$) $\downarrow$
        & $0.59 \pm 0.73$
        & $0.60 \pm 0.60$
        & $0.13 \pm 0.26$
        & $\mathbf{0.02 \pm 0.06}$ \\

        & CD-s ($m^2$) $\downarrow$
        & $0.07 \pm 0.18$
        & $0.17 \pm 0.44$
        & $0.06 \pm 0.19$
        & $\mathbf{0.02 \pm 0.04}$ \\

        \midrule

        \multirow{3}{*}{Revolute}
        & Axis (rad) $\downarrow$
        & $0.82 \pm 0.79$
        & $1.16 \pm 0.52$
        & $0.32 \pm 0.56$
        & $\mathbf{0.08 \pm 0.11}$ \\

        & Position (m) $\downarrow$
        & $0.81 \pm 0.40$
        & $1.18 \pm 1.21$
        & $0.13 \pm 0.25$
        & $\mathbf{0.05 \pm 0.06}$ \\

        & State (rad) $\downarrow$
        & N/A
        & $1.05 \pm 0.59$
        & $0.25 \pm 0.46$
        & $\mathbf{0.15 \pm 0.26}$ \\

        \midrule

        \multirow{2}{*}{Prismatic}
        & Axis (rad) $\downarrow$
        & $0.92 \pm 0.78$
        & $1.27 \pm 0.44$
        & $0.24 \pm 0.33$
        & $\mathbf{0.07 \pm 0.09}$ \\

        & State (m) $\downarrow$
        & N/A
        & $0.63 \pm 0.41$
        & $0.08 \pm 0.22$
        & $\mathbf{0.04 \pm 0.03}$ \\

        \bottomrule
    \end{tabular}
    \label{tab:itaco}
    \vspace{-0.8em}   
\end{table*}

\subsection{Method Study}

A natural first question is how much of the final performance comes from our agentic optimisation  rather than from the underlying VLM alone. We therefore ablate the main components of our harness in \cref{tab:ablation}. A pure agent given the same inputs, task description,
and output format, but without our harness (\textbf{No Harness}), performs substantially worse than the full system.

Providing the agent only with our numerical IoU score (\textbf{IoU Objective}) tooling is not sufficient either: instead, the agent
frequently exploits the objective by producing flat, silhouette-like geometry that achieves high IoU without faithfully reconstructing the
object, resulting in extremely high reconstruction and tracking errors.

We next isolate two components of the optimisation procedure. To evaluate
the importance of VLM-guided pose search, we replace it with differential
evolution over the full pose space while providing ground-truth geometry (\textbf{GT mesh + VLM-free opt.}). Despite this substantial advantage, the variant performs considerably worse than our full method, highlighting the
importance of VLM guidance in identifying useful pose-search regions. Removing the temporal reasoning tool (\textbf{No temporal}) also degrades 3D point tracking, demonstrating the benefit of sequence-level reasoning.

\begin{wrapfigure}[15]{r}{0.50\columnwidth}
    \vspace{-\intextsep}   
    \centering
    \caption{\textbf{Agentic optimisation timelapse on ARCTIC.}
    Tracking error decreases over the course of optimisation.
    The dashed line denotes the no-harness baseline.}
    \includegraphics[
        width=\linewidth
    ]{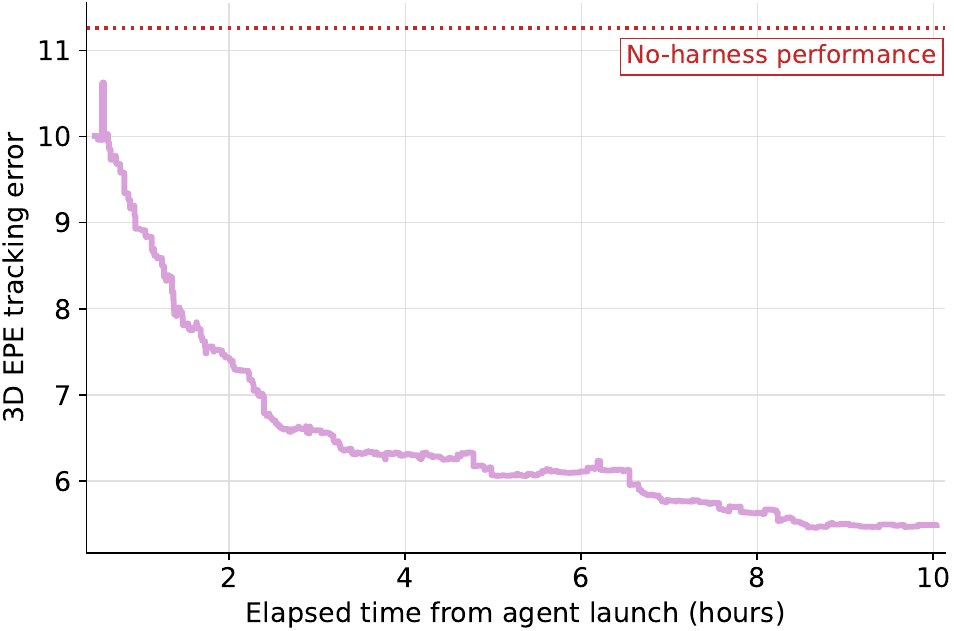}
    \label{fig:arctic_timelapse}
    \vspace{-1.0em}
\end{wrapfigure}

\paragraph{Optimisation dynamics and robustness.}

In~\cref{fig:arctic_timelapse}, tracking error decreases steadily over optimisation, showing convergence behaviour similar to conventional iterative optimisation. As agentic pipelines are typically stochastic, we also report standard deviation of our tracking 3D EPE on the whole dataset obtained over 3 independent runs: $0.06$ cm. Lastly, we swap the underlying agent model to Claude Fable 5 with the same tools and Claude Code harness (\textbf{Fable 5}).

\subsection{iTACO}
Following iTACO~\citep{peng2025itaco}, we evaluate our method on its simulated
RGB-D benchmark using the same keyframes and evaluation protocol. For this
experiment, we use the RGB-D variant of our method and augment the scoring
objective in~\eqref{eq:score} with an $l_1$ depth term. After optimisation,
we scale-align the reconstructed meshes to the input depth maps.

As shown in~\cref{tab:itaco}, our method substantially outperforms existing
baselines on kinematic estimation, while remaining competitive on geometry.
In particular, we achieve the best results on all joint-axis, joint-position,
and joint-state metrics, as well as on two of the three geometry metrics. The baseline methods’ results are taken from iTACO. In $2$ of $73$ sequences (storage furniture objects), our method predicts an additional moving kinematic axis that is absent from the ground truth; this mode is not captured by
the benchmark metrics.

\section{Conclusion and Limitations}
We present AgentSTAR, an approach for shape tracking and reconstruction via agentic optimisation. By combining the coarse visual and structural reasoning of VLMs with numerical optimisation tools, AgentSTAR can recover objects with complex kinematics and track them under challenging conditions that remain difficult for feed-forward systems, including low visual overlap, severe occlusion, and rapid motion. The main limitation of our approach is its inference-time computational cost. However, systems such as AgentSTAR could serve as compute-intensive teachers for the next generation of feed-forward perception models, generating structured reconstructions and trajectories for training. More broadly, alternating between agentic optimisation and feed-forward learning could provide a path toward iterative self-improvement.





\newpage
\bibliography{robotvision}
\bibliographystyle{iclr2026_conference}

\newpage
\appendix
\section{Appendix}

\subsection{Tool List}

We list major utils implemented in our harness and omit infrastructure related tools:
\begin{itemize}
    \item Blender parallel rendering utils coupled with rendering visualisation scripts;
    \item Silhouette scoring, see \cref{sec:scoring};
    \item Pose tool, see \cref{subsec:pose};
    \item Temporal residuals tool, see \cref{sec:temporal};
    \item Turnable: a utility to render views to a sphere around the object.  
    \item External VLM critic invocation for shape quality inspection;
\end{itemize}

\subsection{Sub-agent parallelisation}
Visual inspection for pose estimation creates a computational bottleneck because image inspection is computationally expensive and most agents can inspect only a limited number of images. Image inspection is typically more expensive than the other operations, namely tool invocation and reasoning.

Shape and kinematic structure are centralised decisions. They depend on all frames. Once the shape is coherent enough (so that the notion of pose \textit{makes sense}), we argue that pose estimates are parallelisable.

We adopt a simple schema for motion estimation parallelisation: we split the sequence into $M$ non-overlapping subsequences and spawn a pose \textit{subagent} for each; its sole task is pose estimation for its chunk of ownership. Temporal inconsistencies between the seams are handled by the orchestrator via the same temporal residual tool. Each pose subagent receives a free-form brief indicating whether the current poses are in the correct basin or, e.g., require a flip.

\subsection{Rotation Representation}
We parameterise rotational increments using an ordered composition of rotations around the camera-frame axes for VLM interpretability, since VLM agents excel at reasoning in terms such as ``the object has to be slightly tilted to the left''. Let $\mathbf e_x,\mathbf e_y,\mathbf e_z$ denote the camera-frame basis vectors. Given incremental angles $(\alpha,\beta,\gamma)$, we define:
\begin{equation}
\Delta R(\alpha,\beta,\gamma)
=
\exp\!\left(\alpha[\mathbf e_z]_{\times}\right)
\exp\!\left(\beta[\mathbf e_y]_{\times}\right)
\exp\!\left(\gamma[\mathbf e_x]_{\times}\right).
\end{equation}
Thus, $\Delta R$ is parameterised by successive rotations about the $Z$, $Y$, and $X$ axes of the camera frame.
\subsection{Temporal Residual Derivations}

  Our pose maps canonical object coordinates into camera coordinates:
  \begin{equation}
  \pp_i^{camera}=sR_i\pp+t_i,\qquad t_i=s\tau_i,
  \end{equation}
  where the scale $s$ is shared across frames. Using the camera-to-world
  pose $(\Rcam{i},\tcam{i})$, \cref{eq:x_world} gives:
  \begin{equation}
  \pp^{world}_i
  = s(\Rcam{i}R_i)\pp
  + \left(s\Rcam{i}\tau_i+\tcam{i}\right).
  \end{equation}

  Our temporal tool defines the translational and rotation components of the object's world pose as:
  \begin{equation}
  W_i=\Rcam{i}R_i,\qquad
  u_i=\Rcam{i}\tau_i.
  \end{equation}
  Here, $W_i$ is the object's world orientation, while $u_i$ is the
  camera-to-object offset expressed in world-oriented axes and object
  units. We exclude $\tcam{i}$ because the camera tracker's translation
  gauge may not be compatible with the reconstruction scale.

  For the step from frame $i-1$ to frame $i$, we define angular and translational velocities as:
  \begin{equation}
  \Omega_i=W_{i-1}^{\top}W_i,\qquad
  v_i=u_i-u_{i-1}.
  \end{equation}
  The reported rotational velocity is the geodesic rotation angle
  $r_i^{v}
  =\left\|\log(\Omega_i)\right\|$ 
  expressed in degrees, while the reported translational velocity is $v_i^{t}=\left\|v_i\right\|$ expressed in object units per step.

  At frame $i$, translational acceleration is therefore:
  \begin{equation}
  a_i=v_{i+1}-v_i
     =u_{i+1}-2u_i+u_{i-1},
  \end{equation}
  
  Rotational acceleration is defined as:
  \begin{equation}
  r_i^{a}
  =
   \left\|\log\left(\Omega_i^{\top}\Omega_{i+1}\right)\right\|.
  \end{equation}
\paragraph{Radial Acceleration and Velocity.}
 Radial velocity and acceleration measure changes in the object's depth, which is typically the least constrained direction for monocular methods.
  They are computed along the camera-to-object direction
  $\hat{u}_i=u_i/\|u_i\|$. The signed radial velocity and acceleration are
  $v_i^{\mathrm{rad}}=\hat{u}_i^\top v_i$ and
  $a_i^{\mathrm{rad}}=\hat{u}_i^\top a_i$, respectively. Positive values
  indicate motion or acceleration away from the camera.

  \paragraph{Joints.}
  For joint $m$ with state $q_i^m$, the tool reports joint velocity
  $\Delta_i^m=q_i^m-q_{i-1}^m$ and joint acceleration $A_i^m=\Delta_{i+1}^m-\Delta_i^m$.
  Joint quantities are expressed in their native units: degrees for
  revolute joints and canonical object units for prismatic joints.

  All quantities are reported as differences per step. When camera poses are unavailable, base-pose rotation and translation are computed directly in the camera frames.

\subsection{Scale Alignment}
\label{ap:scale}
Monocular reconstruction is generally recovered up to an unknown global scale.
  Independently moving objects may have a separate scale gauge, even when the
  static scene is metric.

  For each sequence chunk, we estimate one shared scale factor as:
  \[
  \hat{s}=\exp\!\left(
  \operatorname*{median}
  [\log D_{\mathrm{gt}}^t(\mathbf{u})-\log D_{\mathrm{pred}}^t(\mathbf{u})]
  \right),
  \]
  over pixels where both depths are valid. Pooling all frames and using the median makes the estimate robust to outliers. We apply the same alignment to all methods that do not receive metric depth as input.
  
\subsection{Gauge Alignment for HOT3D}
\paragraph{Model free $\SIM$ Gauge alignment.} For model-free 3D object tracking there's a natural $\SIM$ gauge for all predicted estimates, which corresponds to the choice of canonical object coordinate frame with respect to which the object is tracked. Since methods do not observe depth, the scale of the reconstruction and trajectory is not well constrained and is aligned for all methods. 
  \begin{equation}
      p_{\mathrm{gt}} = s G p_{\mathrm{pred}} + c,
      \qquad G \in \SO,\quad c \in \mathbb{R}^3,\quad s>0.
  \end{equation}
  Here, $s$ converts the arbitrary predicted length unit to metric units.

  The predicted and ground-truth object-to-camera mappings are
  \begin{equation}
      q_{\mathrm{pred}}
        = R_{\mathrm{pred}}p_{\mathrm{pred}}+t_{\mathrm{pred}},
      \qquad
      q_{\mathrm{gt}}
        = R_{\mathrm{gt}}p_{\mathrm{gt}}+t_{\mathrm{gt}}.
  \end{equation}
  Because both the predicted geometry and camera-space translation are expressed
  in the same arbitrary unit, the complete predicted camera-space point must be
  scaled by $s$. Thus,
  \begin{align}
      s\left(R_{\mathrm{pred}}p_{\mathrm{pred}} + t_{\mathrm{pred}}\right)
      &=
      R_{\mathrm{gt}}\left(sGp_{\mathrm{pred}} + c\right) + t_{\mathrm{gt}} \\
      &=
      sR_{\mathrm{gt}}Gp_{\mathrm{pred}}
      +R_{\mathrm{gt}}c+t_{\mathrm{gt}}.
  \end{align}
  Equating the linear and translation terms gives:
  \begin{equation}
      R_{\mathrm{pred}} \simeq R_{\mathrm{gt}}G,
      \qquad
      s t_{\mathrm{pred}}
      \simeq t_{\mathrm{gt}}+R_{\mathrm{gt}}c.
  \end{equation}

  We estimate the canonical rotation gauge using $\SO$ Procrustes:
  \begin{equation}
      G^\star =
      \arg\min_{G\in\SO}
      \sum_i
      \left\|R_{\mathrm{pred},i}-R_{\mathrm{gt},i}G\right\|_F^2.
  \end{equation}
  For a fixed scale, the canonical translation is estimated by:
  \begin{equation}
      c^\star =
      \arg\min_c
      \sum_i
      \left\|
        s t_{\mathrm{pred},i}
        -t_{\mathrm{gt},i}
        -R_{\mathrm{gt},i}c
      \right\|_2^2.
  \end{equation}
  While the scaling factor $s$ can also be jointly solved in the same optimisation problem, for methods that also predict a 3D geometric model $s$ is estimated from rendered predicted depth and metric capture depth alignment.

  Therefore, the aligned gauge-aligned trajectory prediction is: 
  \begin{equation}
      \hat R_{\mathrm{pred}} = R_{\mathrm{pred}}{G^{\star}}^{\mathsf T},
      \qquad
      \hat t_{\mathrm{pred}}
        = s t_{\mathrm{pred}}-\hat R_{\mathrm{pred}}c^{\star}.
  \end{equation}

\paragraph{Model-based baselines.} For model-based baselines, such as FoundationPose that consume the metric GT CAD model as input, the gauge ambiguity is not present. However, to ensure the most fair evaluation we align the trajectory if scene object exhibits a natural symmetry (the group of symmetries is provided by the HOT3D dataset natively). We select the symmetry element minimising the rotational error in the first visible frame and apply this same element to every predicted pose. The selected symmetry is fixed across the entire sequence, and no additional pose alignment is performed.

\section{Dataset Use Statement}
The ARCTIC~\citep{fan2023arctic}, HOT3D~\citep{banerjee2025hot3d}, and PartNet-Mobility~\citep{peng2025itaco} datasets were used in this work solely for scientific research purposes. Their use was limited to the benchmarking and evaluation described in this paper and its supplementary material.

\end{document}